\documentclass[runningheads]{llncs}
\usepackage{graphicx}
\usepackage{amsmath,amssymb} 
\usepackage{color}
\usepackage{subfigure}
\usepackage{epstopdf}
\usepackage{multirow}
\usepackage{indentfirst}
\usepackage{tabularx, booktabs}
\usepackage{stackengine}
\usepackage{url}
\newcolumntype{Y}{>{\centering\arraybackslash}X}
\newcommand{\etal}{\textit{et al}.}
\usepackage[width=122mm,left=12mm,paperwidth=146mm,height=193mm,top=12mm,paperheight=217mm]{geometry}
\begin{document}
\pagestyle{headings}
\mainmatter

\title{Exploring Local Context for Multi-target Tracking in Wide Area Aerial Surveillance} 
\author{Bor-Jeng Chen \and
G\'erard Medioni
}
\institute{University of Southern California\\ {\tt\small \{borjengc, medioni\}@usc.edu}}

\maketitle

\begin{abstract}
Tracking many vehicles in wide coverage aerial imagery is crucial for understanding events in a large field of view. Most approaches aim to associate detections from frame differencing into tracks. However, slow or stopped vehicles result in long-term missing detections and further cause tracking discontinuities. Relying merely on appearance clue to recover missing detections is difficult as targets are extremely small and in grayscale. In this paper, we address the limitations of detection association methods by coupling it with a local context tracker (LCT), which does not rely on motion detections. On one hand, our LCT learns neighboring spatial relation and tracks each target in consecutive frames using graph optimization. It takes the advantage of context constraints to avoid drifting to nearby targets. We generate hypotheses from sparse and dense flow efficiently to keep solutions tractable. On the other hand, we use detection association strategy to extract short tracks in batch processing. We explicitly handle merged detections by generating additional hypotheses from them. Our evaluation on wide area aerial imagery sequences shows significant improvement over state-of-the-art methods.
\keywords{multi-target tracking, context tracker, wide area motion imagery}
\end{abstract}

\section{Introduction}
\label{sec:intro}
Wide area motion imagery (WAMI) is acquired by high altitude unmanned aerial vehicles (UAV) and has made it possible to understand activities in a large area of interest. With current sensor and storage technologies, WAMI is typically captured in large format (tens $\sim$ hundreds of megapixels), low frame rate (1 $\sim$ 2 Hz), grayscale, and with ground sampling distance from 0.2 $\sim$ 0.5 meter/pixel. Because of these unique characteristics and its wide applications, it has gained attention in the computer vision field in recent years \cite{Reilly2013,ChenWACV15,Liao2011,Kang2015,ShuCVPR15,ProkajCVPR14,ChenAVSS15}. Tracking many vehicles in WAMI \cite{ProkajCVPR14,ChenAVSS15,Perera2006,Reilly:2010,Saleemi2013} is an essential component since it is needed for higher level activity analysis and scene understanding. 

\begin{figure}[t]
\begin{center}
\subfigure{
\includegraphics[width=0.6\linewidth]{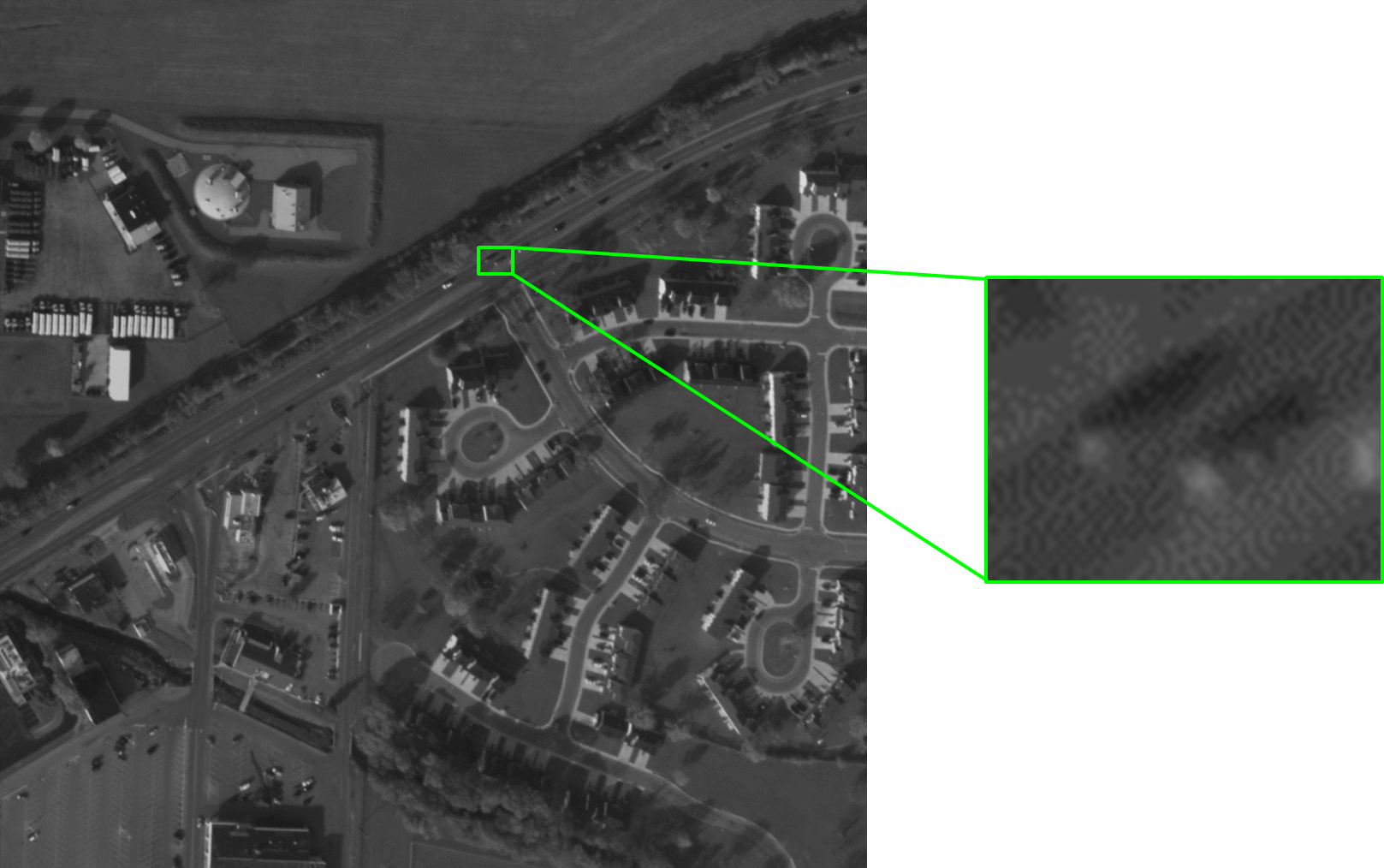}}
\end{center}
\caption{A cropped view of WPFAB dataset \cite{WPAFB:Online}. The green rectangle shows details of two vehicles which are close to each other.}
\label{fig:samples}
\end{figure}

Detection association strategy has become standard for multi-target tracking with unknown number of targets \cite{Andriluka2008,Breitenstein2009,Collins2012}. By assigning detections in each frame into tracks, short-term missing detections are recovered by using motion interpolation. In the following, we call this type of tracker detection-based tracker (DBT). Unique characteristics in WAMI bring additional challenges in DBT. First, low frame rate WAMI leads to a very large search space. Displacement of a target is up to 80 pixels in consecutive frames. Second, the size of a vehicle is extremely small (usually less than 20 pixels in length). Along with grayscale imagery, appearance information is less discriminating in WAMI than in many other scenarios. Fig. \ref{fig:samples} shows a cropped view from a WAMI dataset. Notice that in this work, we assume that vehicles are the only moving targets, since other moving objects such as pedestrians are nearly invisible even for human beings. 

Instead of using an appearance-based detector, current tracking approaches in WAMI \cite{ChenAVSS15,Perera2006,Reilly:2010,Saleemi2013} rely on motion detection. They first stabilize the imagery and then apply frame differencing methods. However, stopped or slow vehicles result in long-term missing detections, which can not be recovered using DBT only.  

To alleviate the limitation of detectors, a new strategy that couples DBT with an appearance-based category free tracker (CFT) has been proposed in multi-target tracking \cite{Yang:2012,Wang15}. CFT does not rely on a detector and it recovers missing detections at the ends of a track. However, appearance-based CFT is not robust in WAMI because targets are with weak appearance. Moreover, it is difficult to escape from local maximums when the frame rate is low. Prokaj and Medioni \cite{ProkajCVPR14} proposed to run DBT and a regression based tracker in parallel. Though using a regressor is more efficient than using a classifier in low-frame-rate WAMI, the regression model is still not discriminating enough to avoid drifting. 

Based on the above discussion, incorporating CFT with DBT in WAMI still remains very challenging. We conclude that there are three major issues that make it difficult in WAMI: 1) Discriminating ability of target appearance is very low because of the small target size and grayscale imagery. Illumination change can easily confuse a target appearance model with its local background or other neighboring targets. 2) Motion detections are imprecise especially at merged detections, which contain more than one target. Thus, it is difficult to have good initialization for CFT. 3) Displacement of a target between consecutive frames can be very large. A good strategy for hypothesis sampling in CFT is essential to keep the solution tractable. 

\begin{figure}[t]
\begin{center}
\subfigure{
\includegraphics[width=0.98\linewidth]{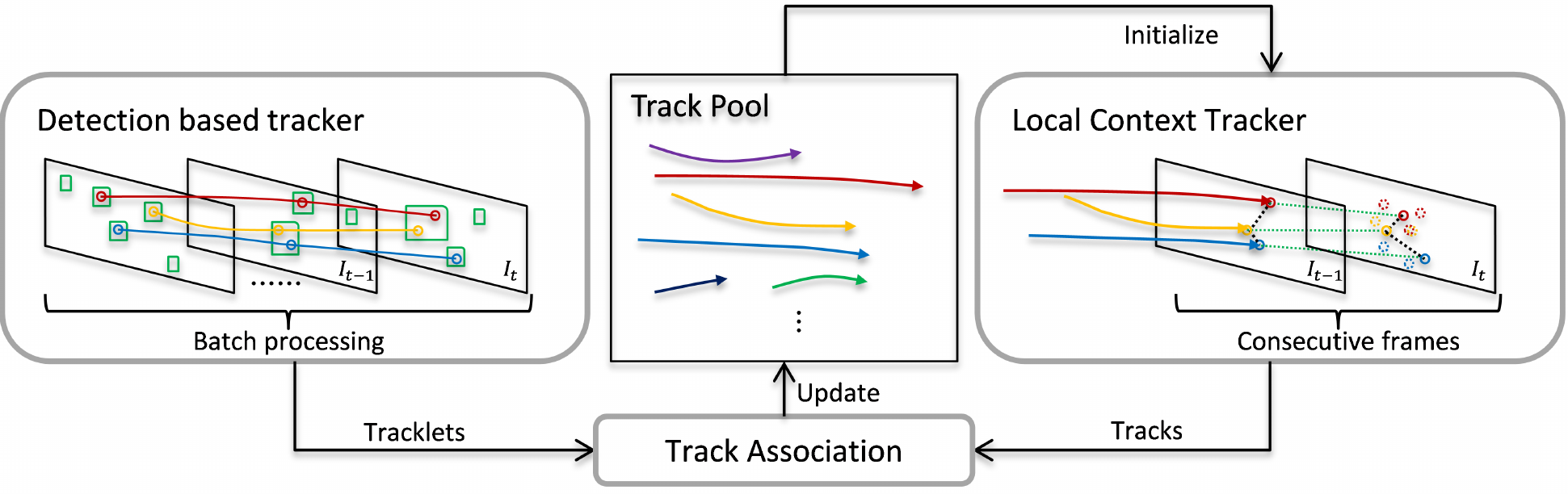}}
\end{center}
\caption{Framework of our tracking system. Green rectangles represent motion detections. Solid circles show hypothesis centers that form tracks. Dashed circles represent hypothesis centers in LCT. Dashed black lines show context constraints in LCT.}
\label{fig:flwchart}
\end{figure}

Our goal is to maximize the merit of both trackers and compensate their limitations in WAMI. Unlike \cite{Yang:2012,Wang15,ProkajCVPR14} training an appearance model for CFT, we relax the dependency on appearance clue, which is not reliable in our scenario, by introducing \textit{local context tracker} (LCT). LCT explores spatial relations for a target to avoid unreasonable model deformation in the next frame. We design two sampling strategies based on dense and sparse optical flow to overcome large search space in low frame rate aerial videos. In DBT, short tracks (tracklets) are produced by associating hypotheses from motion detections in a sliding temporal window. We explicitly handle merged detections by generating additional hypotheses from them. This step is important for combing DBT with LCT to ensure reasonable appearance and motion consistency. The track association step concludes results from both trackers and updates the ``track pool", which stores all existing tracks and is used to initialize LCT in the next frame. Fig. \ref{fig:flwchart} illustrates the framework of our system.

The contributions of this paper are:\\
1. We propose LCT that relaxes the dependency on frame differencing motion detection and appearance information.\\
2. We propose DBT that explicitly handles merged detections in detection association.\\
3. We propose a unified framework that couples LCT with DBT and takes advantages of both trackers.\\
4. Our performance shows significant improvement over state-of-the-art methods in two WAMI sequences.

The rest of the paper is organized as follows: We discuss related work in Section \ref{sec:rel}. We then illustrate LCT in Section \ref{sec:LCT}. Our DBT is introduced in Section \ref{sec:DBT}. The track association module is described in Section \ref{sec:ass}. In Section \ref{sec:exp}, we show comparison results on two sequences from real WAMI datasets. Finally, we conclude this paper in Section \ref{sec:conc}.
\section{Related Work}
\label{sec:rel}
Multi-target tracking has been investigated in the computer vision society for many years. Joint probabilistic data association filter (JPDAF)\cite{jpdaf} and multiple hypothesis tracking (MHT)\cite{mht} are two early successful approaches. However, the association step requires very high computational and memory cost in both methods; therefore, solutions are usually intractable in real-world applications. In practice, JPDAF has been incorporated with Kalman filter \cite{Kang03} and particle filter \cite{JPDAparticle} to increase the efficiency. More recently, Rezatofighi \cite{jpdaR} improves the efficiency of JPDAF by obtaining m-best solution using integer linear programming and shows state-of-the-art result. MHT usually introduces tree pruning strategies \cite{fastMHT,ChenouardPAMI2013,MHTR} to reduce the possible solution space. In recent years, network flow optimization has become popular in multi-target tracking approaches \cite{Kpath,Networkflow,mincost,Pirsiavash11}. Despite that these methods have shown promising results in their scenario, they all use one-to-one matching assumption. Therefore, they are not suitable for WAMI where split-and-merge motion detections often occur.

Solving tracking problem with machine learning techniques has shown to be effective in boosting discriminative ability of appearance model for single target tracking \cite{IVT,MLT,TLD} and multi-target tracking \cite{KuoCVPR10,Yang:2012,xiang_iccv15}. However, it is almost impossible to learn meaningful information in WAMI because the target size is extremely small and the imagery is typically in grayscale. Targets are often visually similar to each other and background patches.

Using context information is an appealing strategy for tracking against distracters and occlusion. This concept has been applied on single target tracking \cite{ContextTracker,InvisibleTrack,ContextAware}. Recently, \cite{SPOT} incorporates spatial constraints with tracking-by-detection. Nevertheless, these trackers require accurate annotation for initialization. It is not trivial to directly apply these approaches in WAMI, where the number of targets varies with time and perfect initialization is not available.  

Most WAMI tracking approaches focus on associating noisy motion detections into tracks. Motion detections are typically acquired by applying frame differencing methods to stabilized imagery. Perera \etal ~\cite{Perera2006} propose to first generate short tracks using nearest-neighbor strategy and then handle split-then-merge situations in track linking. Reilly \etal ~\cite{Reilly:2010} formulate the data association problem in Hungarian algorithm. Prokaj \etal ~\cite{ProkajCVPRW} extract tracklet from detections by Bayesian network. Shi \etal ~\cite{Shi2013} associate motion detections by rank-1 tensor optimization. Keck \etal ~\cite{Keck13} provide a real-time implementation for tracking based on multiple hypothesis tracking. The object-centric association method is proposed in \cite{Saleemi2013} to relax the one-to-one matching assumption for motion detections. Additional context constraints are used to alleviate track intersection. Chen and Medioni \cite{ChenAVSS15} extract tracklets by finding the longest path through detection trees. These above trackers mainly rely on motion detections. Therefore, they cannot recover long-term missing detections from slow or stopped vehicles. 

Xiao \etal ~\cite{Jiangjian10} propose to use appearance and shape templates to handle missing detections. To avoid drifting, they use road network information and consider pairwise spatial relation in optimization. However, road network information is not always available and considering spatial relations in Hungarian optimization is costly. Basharat \etal ~\cite{Basharat14} apply an appearance-based tracker whenever detection association fails for a track or the motion of a target is slow. More recently, a hybrid approach that combines DBT with a regression-based tracker is proposed \cite{ProkajCVPR14} to handle stop-then-go vehicles. Nevertheless, relying only on weak appearance information makes these trackers prone to drift and limits its ability in recovering missing detections.

\section{Local Context Tracker}
\label{sec:LCT}
In this section, we present the details of our LCT. It explores context information to increase the robustness of tracker. Given a track pool ${\mathcal T_e}=\{T_1,T_2,...,T_N\}$, which contains $N$ tracks from time $1$ to $t-1$, our goal is to extend each track that ends at the previous frame $I_{t-1}$ to the current frame $I_t$. We introduce our hypothesis generation method using sparse and dense flow in Section \ref{subsec:hypo}. In Section \ref{subsec:opt}, we formulate the tracking problem in graph optimization and find the optimal hypothesis tree considering appearance, motion, and context information. 

\subsection{Hypothesis Sampling}
\label{subsec:hypo}
As mentioned in Section \ref{sec:intro}, the displacement of a target is in a very large range (0$\sim$ 80 pixels). In practice, it is not affordable to densely sample all possibilities. We propose a search mechanism which is based on motion information. We use the fact that visible targets are either fast enough to produce discontinuities in dense flow or slow enough to meet the small motion assumption of optical flow. For each track $T_i$, we construct a set of hypotheses $H_i$ for graph optimization in Section \ref{subsec:opt}.

\begin{figure}[t]
\begin{center}
\subfigure{
\includegraphics[width=0.6\linewidth]{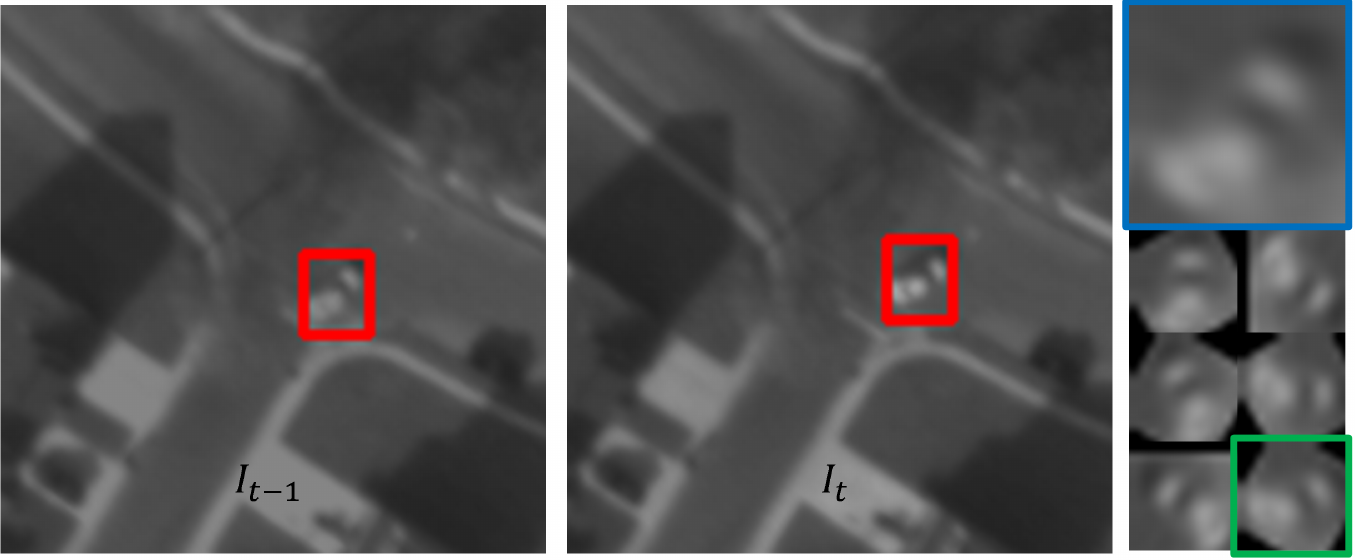}}
\end{center}
\caption{A turning vehicle and rotation variants of $s_i$. Blue rectangle represents $s_i$ and the green rectangle represents the selected template with the highest NCC score.}
\label{fig:rots}
\end{figure}

\subsubsection{Sample from Dense Flow}
\label{subsub:dense}
In WAMI, most of the targets move fast and the small motion assumption fails in optical flow methods. Therefore, motion vectors from optical flow in a target region are usually noisy. Instead of using the motion vectors directly, we find discontinuities in dense flow for hypothesis sampling. We use $3 \times 3$ Sobel operators to get the gradient of flow and produce a binary voting map by thresholding the gradient magnitude. We search each target in a $160 \times 160$ window based on its location and size at previous frame. Here, we assume that the target size is consistent between consecutive frames because roads are typically flat and the UAV does not change its altitude drastically while flying. A valid hypothesis should have enough votes from the binary voting map. We set the threshold as one-fourth of the target size in all experiments.

\begin{figure}[t]
\begin{center}
\subfigure[target at $I_{t-1}$]{
\includegraphics[width=0.25\linewidth]{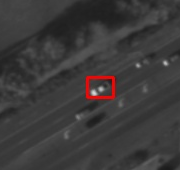}}
\subfigure[binary voting map]{
\includegraphics[width=0.25\linewidth]{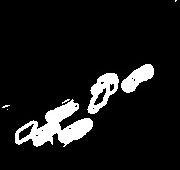}}
\subfigure[hypotheses at $I_t$]{
\includegraphics[width=0.25\linewidth]{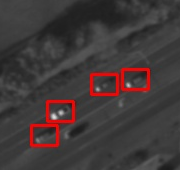}}
\end{center}
\caption{An example of generating hypotheses using a binary voting map.}
\label{fig:denseflowhypo}
\end{figure}
In addition, we ensure that hypotheses match the target by using two kinds of templates including the target region at the previous frame $o^i_{t-1}$ and a ``stable template" $s_{i}$ for the track $T_i$.  We use normalized correlation coefficient (NCC) as template matching score. Non-maximum suppression is applied to reduce the number of candidates. The idea of stable template is to maintain a robust appearance model that only updates at high confidence to avoid gradually drifting \cite{TMP}. In this work, we take the advantage of DBT for the confidence measure. We describe the update criterion of stable template in Section \ref{sec:ass}. 

NCC template matching handles target motion in translation but not in rotation. Therefore, it fails when a target turns if we do not consider rotation of $s_i$, which may have different orientation compared with the target at the current frame. A collection of rotation variants of $s_{i}$ are used so that the tracker can deal with turning targets as shown in Fig. \ref{fig:rots}. We use 7 rotation variants  ($-90$, $-60$, $-30$, $0$, $30$, $60$, $90$ degrees) in our implementation. The following rules are used to adopt hypotheses with strong appearance similarity:
\begin{equation}
NCC(h_t,o^i_{t-1})> \phi,
\label{eq:valhypo}
\end{equation}
\begin{equation}
MAX_j(NCC(h_t,s^j_{i}))> \phi, 
\label{eq:valChypo}
\end{equation}
where $\phi$ is a constant which is set to 0.5, $h_t$ represents a template of a hypothesis candidate, and $s^j_{i}$ is the $j_{th}$ rotation variant of $s_{i}$. Fig. \ref{fig:denseflowhypo}(c) shows an example of samples from the dense flow. Note that generating hypotheses directly from connected components of the voting map is not suitable since merged blobs appear frequently in practice. Fig. \ref{fig:denseflowhypo}(b) shows an example of this situation.

\subsubsection{Sample from Sparse Flow}
When a target moves slowly or stops, there is no discontinuity in dense flow between the target region and its local background region. Fortunately, small motion assumption of optical flow is valid in this case. We attempt to generate a hypothesis from densely sampled sparse flow to handle the situation that a target moves relatively slow or stops. For each pixel in the target bounding box at the previous frame, we use Lucas-Kanade optical flow \cite{BouguetLK} to track them at the current frame. Note that target hypotheses with fast motion are already covered by using dense flow in the previous section. Thus, we only use a relatively small window for sparse flow. In all our experiments, this window is set to $15 \times 15$ in pixels. 

\begin{figure}[t]
\begin{center}
\subfigure[Slow target]{
\includegraphics[width=0.45\linewidth]{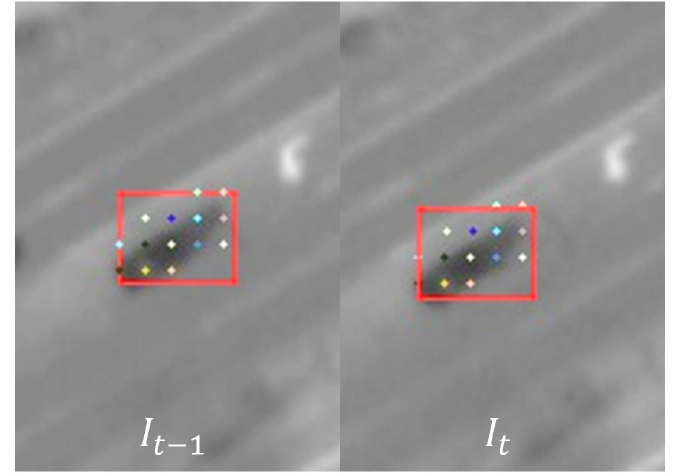}}
\subfigure[Occluded target]{
\includegraphics[width=0.45\linewidth]{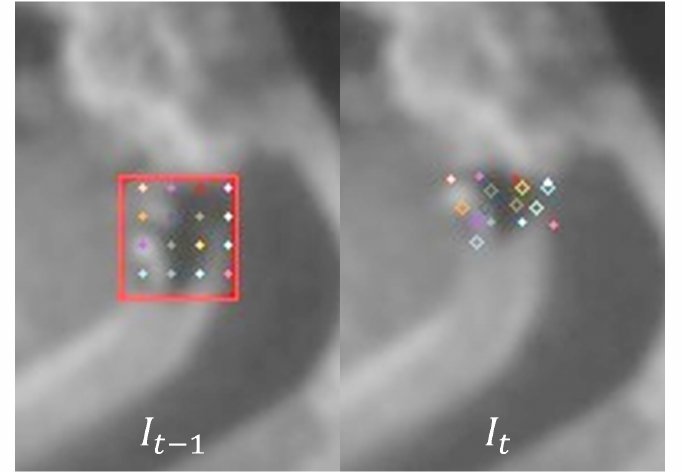}}
\end{center}
\caption{Different situations of sampling from the sparse flow. Points with the same color represent correspondences from the sparse optical flow. Hollow rhombuses show failure points in forward-backward check. (a) Optical flow tracks most points successfully when the target is slow and visible. A valid hypothesis in $I_t$ is shown in the red rectangle. (b) Most points fail in forward-backward check when the target is occluded. No hypothesis is generated in this case.}
\label{fig:sparseflowhypo}
\end{figure}

The main reason of using sparse flow is that we can apply a forward-backward check \cite{TLD,xiang_iccv15} efficiently to remove inconsistent correspondences. The median flow vector of remaining points is used to predict the hypothesis region. We avoid introducing false alarms by rejecting hypotheses without enough number of valid correspondences as shown in Fig. \ref{fig:sparseflowhypo} (b). The threshold is set to $1/8$ of the target size at $t-1$. Additionally, two NCC template matching policies, which are illustrated in equation \ref{eq:valhypo} and \ref{eq:valChypo}, are applied to select valid hypotheses. 

\subsection{Optimization in Hypothesis Graph}
\label{subsec:opt}
\begin{figure}[t]
\begin{center}
\subfigure[]{
\includegraphics[width=0.25\linewidth]{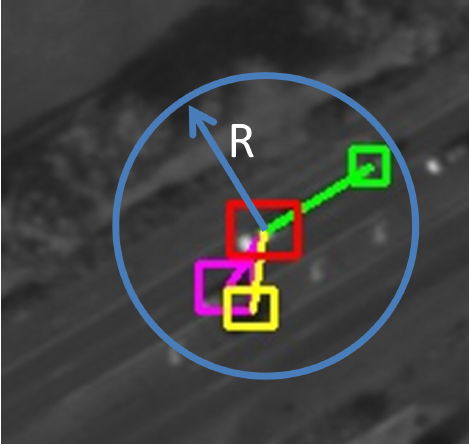}}
\subfigure[]{
\includegraphics[width=0.35\linewidth]{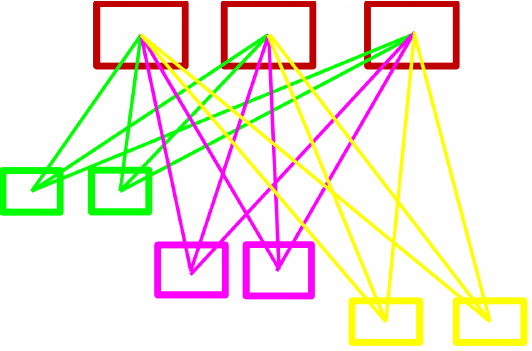}}
\subfigure[]{
\includegraphics[width=0.25\linewidth]{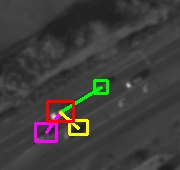}}
\end{center}
\caption{(a) Context relation at $I_{t-1}$ (b) Hypothesis graph at time $t$ (c) The tracking result of LCT at $I_t$}
\label{fig:graph}
\end{figure}

Assuming independence between all tracks causes ambiguities between neighboring targets with similar appearance in WAMI. While considering every linking possibilities between all hypotheses and all tracks is computationally expensive, we argue that each target trajectory is mainly affected by other targets with a similar motion direction in a local neighborhood. The movement of each vehicle is constrained to avoid collisions with its neighbors. For each target track $T_i \in {\mathcal T_e}$, we find its neighboring tracks with similar moving direction at $t-1$ as its ``motion neighbors". We resolve the tracking problem for each $T_i$ and its motion neighbors at the same time. Fig. \ref{fig:graph} (a) shows the neighborhood of the target (red). In our implementation, we use a search radius $R=50$ pixels.

Given a target track $T_i$, its motion neighbors, and their hypothesis sets, we construct a hypothesis graph $G = (V,E)$ as shown in Fig. \ref{fig:graph} (b). Each sample in a hypothesis set is represented by a node, while edges are constructed between every node from target track and every node from each of its motion neighbor. Let $k=\{{k^i_t|^{|V|}_{i=1}, k^i_t \in V}\}$ be a set of nodes that forms a tree, we formulate the tracking problem as maximizing the following objective function:
\begin{equation}
O(k)=\lambda\sum\limits_{k^i_t\in V} U(k^i_t)+\sum\limits_{(k^i_t,k^j_t)\in E} B(k^i_t,k^j_t),
\label{eq:enegry}
\end{equation}
where $U(k^i_t)$ is the unary score at node $k^i_t$, and $B(k^i_t,k^j_t)$ is the binary score between nodes that forms an edge in $E$. We use $\lambda$, which is set to $3$ in our experiments to leverage weighting between these two scores. Using the trajectory of $T_i$, we define $U(k^i_t)$ as:
\begin{equation}
U(k^i_t)=C_{app}(k^i_t)\cdot C_{mot}(k^i_t),
\label{eq:unary}
\end{equation}
where $C_{app}(k^i_t)$ is the appearance measurement based on NCC calculated between the template of hypothesis and the template of its corresponding target at the previous frame. $C_{mot}(k^i_t)$ is computed by multiplying the velocity similarity with the acceleration similarity. Here, we adopt the same velocity and acceleration measure as in \cite{ChenAVSS15}, which uses different Gaussian kernels for magnitude and orientation components.   
\begin{figure}[t]
\begin{center}
\subfigure{
\includegraphics[width=0.48\linewidth]{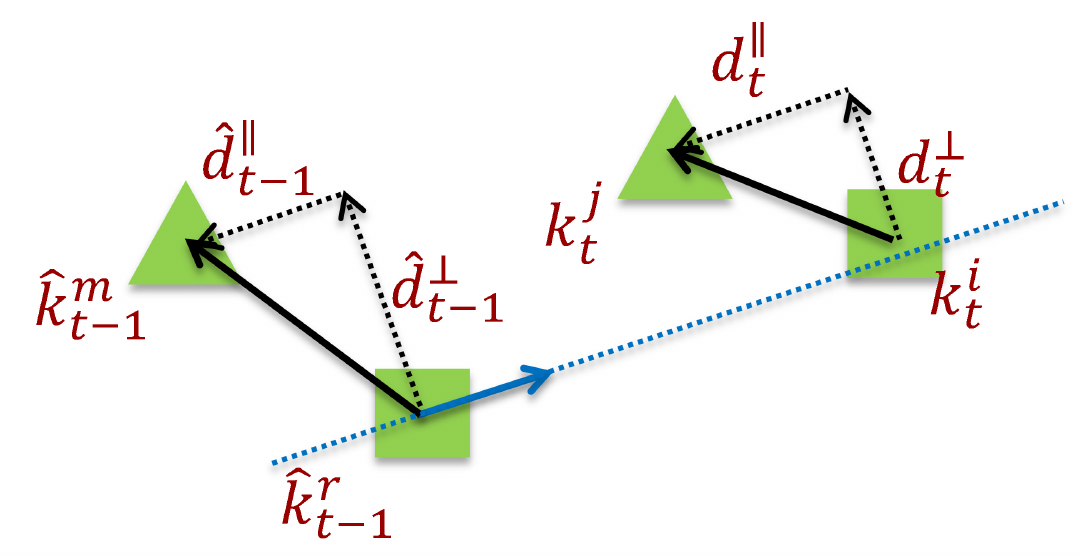}}
\end{center}
\caption{Context constraint model at time $t-1$ and one of the hypothesis pairs at time $t$. Triangles represent the motion neighbor and its hypothesis while square represents the target and its hypothesis. The blue arrow shows the latest velocity vector of the target.}
\label{fig:context}
\end{figure}

We calculate a binary score $B(k^i_t,k^j_t)$ according to the local context. This provides discriminating power against id-switches and false alarms. Given a target track and one of its motion neighbors, we observe that the relative displacement between them is more flexible along the moving direction of the target than along its normal direction. Therefore, we decompose the displacement according to the target velocity vector and use different Gaussian kernels for structure consistency in different components. Fig. \ref{fig:context} shows our local context modeling. $\hat{d}^{\parallel}_{t-1}$ and $\hat{d}^{\bot}_{t-1}$ are spatial relations between the target and the neighbor while $d^{\parallel}_t$ and $d^{\bot}_t$ represent one of the hypothesis pairs between them. The binary score is defined as:
\begin{equation}
B(k^i_t,k^j_t)=C_{\parallel}(k^i_t,k^j_t)\cdot C_{\bot}(k^i_t,k^j_t),
\label{eq:binary}
\end{equation}
\begin{equation}
C_{\parallel}(k^i_t,k^j_t)=exp(-\alpha\cdot Abs(\hat{d}^{\parallel}_{t-1}-d^{\parallel}_t)),
\label{eq:binary}
\end{equation}
\begin{equation}
C_{\bot}(k^i_t,k^j_t)=exp(-\beta\cdot Abs(\hat{d}^{\bot}_{t-1}-d^{\bot}_t)),
\label{eq:binary}
\end{equation}
where $Abs(\cdot)$ returns the absolute value. We set $\alpha=0.01$ and $\beta=0.05$  as constants in our experiment to penalize deformation in different components.

The optimization can be solved by using dynamic programming efficiently in polynomial time \cite{DP11}. Fig. \ref{fig:graph} (c) shows an example LCT tracking result. By considering the local context as well as the trajectory history, LCT does not drift to other targets or false alarms.

\section{Detection Based Tracker}
\label{sec:DBT}
One of the difficulties in coupling DBT and LCT is: LCT assumes that the observation at the end of a track contains only one target. This assumption fails when a merged detection occurs. The appearance of these detections accounts for multiple targets. The center of them may be far from any target. This makes both appearance and motion information unreliable for LCT initialization. Unlike most DBT methods that do not handle this situation in the detection association level, we address this problem by generating additional hypotheses rather than merely adopt motion detections directly.

Our DBT is based on the motion proprogation approach proposed in \cite{ChenAVSS15}. It builds detection trees layer-by-layer from motion detections at each frame in a sliding temporal window, which is set to 8 frames in our implementation. Each tree node represents a detection. Edges are constructed iteratively based on the best path from the root to each node at the previous frame. Each detection tree produces at most one tracklet by finding an optimal path from root to leaf based on appearance and motion consistency. 

\begin{figure}[t]
\begin{center}
\subfigure[]{
\includegraphics[width=0.25\linewidth]{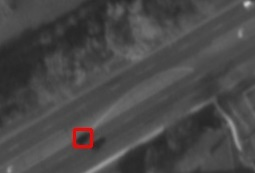}}
\subfigure[]{
\includegraphics[width=0.25\linewidth]{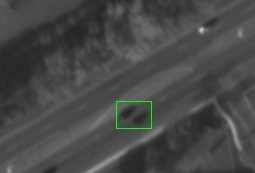}}
\subfigure[]{
\includegraphics[width=0.25\linewidth]{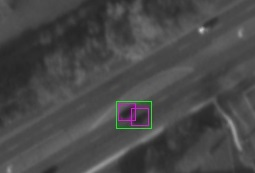}}
\end{center}
\caption{Generate additional nodes from merged detection. (a) parent node (red) (b) child node (a merged detection)  (green) (c) child node with additional nodes (purple).}
\label{fig:merge}
\end{figure}

Using this framework, we identify abnormal changes of detection size for each edge in a detection tree. Then, we estimate additional hypotheses from these cases to improve tracking accuracy. Let $d_{t-1}$ be a parent detection of a child node, $d_t$. If $1.5\cdot Size(d_{t-1})<Size(d_t)$, we call $d_t$ a ``potential merged detection". $Size(d)$ returns the size of a detection $d$. Fig. \ref{fig:merge} (a) shows an example of $d_{t-1}$ and (b) shows an example of $d_t$.

Appearance consistency is used to generate hypotheses from a potential merged detection. We scan the region with the template of the parent node and calculate NCC score. Local maximums with scores larger than $\phi$ produce new tree nodes. We insert the new nodes into the detection tree and add a link between each new node and $d_{t-1}$ to update the detection tree. In Fig. \ref{fig:merge} (c), purple rectangles show two additional nodes from a potential merged detection in the green rectangle. Their parent node is shown in Fig. \ref{fig:merge} (a) in the red rectangle. By inserting these hypotheses, the optimal solution avoids choosing merged detections. This is because the additional hypothesis at the correct location leads to higher appearance and motion consistency than a merged detection, which is suboptimal.

\section{Track Association}
\label{sec:ass}
With new tracklets from DBT and tracking results from LCT, we describe how to generate the final result in this section. This result is further used to update the track pool. We find associations between new tracklets ${\mathcal T_n}$ from DBT and existing tracks ${\mathcal T_e}$ in the track pool in two steps. Let $t^n_{st}$, $t^n_{ed}$, $t^e_{st}$, $t^e_{ed}$ be the first frame index and the end frame index of a new tracklet $T_n$ and an existing track $T_e$. We define the association score $A(T_e,T_n) = S_p(T_e,T_n) \times S_v(T_e,T_n)$, where $S_p(T_e,T_n)$ and $S_v(T_e,T_n)$ represents the position and velocity similarity. Since most successfully associated pairs are with overlapping time, given $T_n$, the first step is to find the best match among existing tracks with $t^e_{st} \geq t^n_{ed}$. In this case, we calculate $S_p(T_e,T_n)$ as:
\begin{equation}
S_p(T_e,T_n)=Match(T_e,T_n)/(t^e_{st}-t^n_{ed}+1),
\label{eq:sp}
\end{equation}
where $Match(T_e,T_n)$ returns the number of matched observation pairs in the overlapping period. Here, we define an observation as a bounding box region in a track or a tracklet. Given a time index, if the center of an observation from a tracklet lies in the observation from an existing track or vice versa, we treat them as matched observation. $S_v(T_e,T_n)$ is based on the Euclidean distance between the velocities from $T_e$ and $T_n$ at time $t^n_{st}$ with a Gaussian kernel. The constant threshold $\zeta=0.6$ is used to adopt valid association.

If the successful association cannot be found in overlapping existing tracks, we interpolate the first observation of $T_n$ to time $t^e_{ed}$ of each non-overlapping track using a linear motion model and compute $S_p(T_e,T_n)=exp(-0.01*Dist(T_e,T_n))$, where $Dist(T_e,T_n)$ is the Euclidean distance between the interpolated observation and the last observation of $T_e$. $S_v(T_e,T_n)$ is calculated using the Euclidean distance between velocities at $t^n_{st}$ in $T_n$ and at $t^e_{ed}$ in $T_e$. Again, $\zeta$ is used to accept successful association pairs. Unassociated tracklets initialize new tracks in the track pool.

Given a track in the track pool, if a tracklet is associated with it and LCT also tracks the target successfully, we append the observation at time $t$ with larger NCC score compared with the observation at time $t-1$ in the tracklet. If only one of the trackers produce a valid result, we will extend the track using this result. 

Updating template robustly is difficult in tracking problems. A template has to adapt to changes of target appearance. At the same time, the template should not drift to background gradually because of the update. Fortunately, DBT tracklet provides strong evidence in target existence without using template information. Thus, we update the stable template of a track whenever a valid tracklet is associated with it. The latest observation of the track is used to update the rotation variants of the stable template.
\section{Experiments}
\label{sec:exp}
\subsection{Setup}
We compare our methods with state-of-the-art trackers \cite{Reilly:2010,ProkajCVPRW,ProkajCVPR14,jpdaR,ChenAVSS15} on two WAMI sequences. In order to get a fair comparison, we use the same motion detection result as input for all trackers. The detection approach is based on background subtraction with 3-D stabilization \cite{ChenWACV16}, which reduces most false alarms from parallax effect. 

We obtain executables of \cite{Reilly:2010} and \cite{ProkajCVPR14} from the authors of \cite{ProkajCVPR14}. The programs of \cite{ProkajCVPRW,ChenAVSS15} are provided by their authors. We do not change parameters for the above methods except for the metadata information, which includes frame rate and the ground sampling distance of imagery. Evaluation of \cite{jpdaR} is based on the MATLAB code provided by its authors. We tune parameters based on their helpful advice in order to apply it on low-frame-rate WAMI.
\subsection{Evaluation Metrics}
Our quantitative evaluation is based on commonly used metrics which are also adopted in \cite{ProkajCVPR14,ChenAVSS15}, including \textit{recall}: number of true positive detection/number of ground truth detection; \textit{precision}: number of true positive detection/number of detection in tracks; \textit{false positive per frame (FP/F)}; \textit{false positive per ground truth detection(FP/GT)}; \textit{multiple object detection accuracy (MODA)}; \textit{number of swaps (id-switches) per track (S/T)}; \textit{number of breaks per track (B/T)}; \textit{multiple object tracking accuracy (MOTA)}. The definition of \textit{MODA} and \textit{MOTA} can be found in \cite{Eval}.

\subsection{Results on WPAFB 2009 Sequence}
The sequence is selected and preprocessed by the authors of \cite{ProkajCVPR14} from a public WAMI dataset \cite{WPAFB:Online}. This dataset is recorded around 1 Hz and it provides ground truth labels for vehicles. The sequence covers a 429 m $\times$ 429 m suburb area in OH, USA with 1408 pixels $\times$ 1408 pixels. The ground sampling distance is around 0.3 meters per pixel. There are 1025 frames and 410 tracks. Several stop-then-go situation happens in the scene. Many merged detections appear when vehicles are close to each other.
\begin{table*}[t]
\begin{center}
\begin{tabular}{|c|c|c|c|c|c|c|c|c|}
\hline
Method & Recall & Precision & FP/F & FP/GT & MODA & S/T & B/T & MOTA\\
\hline
Reilly \etal~\cite{Reilly:2010} 
& 0.573 & 0.94 & 0.887 & 0.037 & 0.536  & 0.851 & 1.293 & 0.522\\ 
Prokaj \etal~\cite{ProkajCVPRW} 
& 0.504 & 0.985 & 0.18 & 0.007 & 0.497 & 0.249 & 1.515  & 0.493\\
Prokaj \etal~\cite{ProkajCVPR14}   
& 0.539 & 0.96 & 0.548 & 0.023 & 0.516 & 0.237 & 1.022 & 0.512\\
Rezatofighi \etal~\cite{jpdaR}
& 0.44 & 0.74 & 3.746  & 0.155 & 0.285 & 0.529 & 2.855 & 0.276\\
Chen \etal~\cite{ChenAVSS15}
& 0.55 & 0.987 & 0.171 & 0.007 & 0.543  & 0.2 & 0.5 & 0.54\\
\hline
Ours (DBT only) & 0.537 & 0.985 & 0.195 & 0.008 & 0.529 & 0.061 & 0.573 & 0.528\\
Ours (DBT+LCT) & {\bf 0.606} & {\bf 0.99} & {\bf 0.145} & {\bf 0.006} & {\bf 0.6} & {\bf 0.015} & {\bf 0.317} & {\bf 0.599}\\
\hline
\end{tabular}
\end{center}
\caption{Comparison of tracking results of the WPFAB 2009 sequence. Best results in each indexes are shown in bold text.}
\label{tab:one}
\end{table*}

\begin{figure}[t]
\begin{center}
\includegraphics[width=0.15\linewidth]{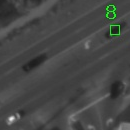}
\includegraphics[width=0.15\linewidth]{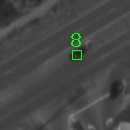}
\includegraphics[width=0.15\linewidth]{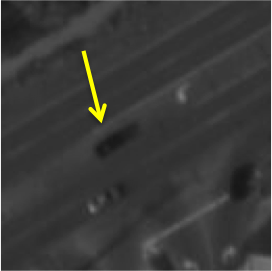}
\includegraphics[width=0.15\linewidth]{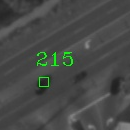}
\includegraphics[width=0.15\linewidth]{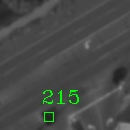}\\
\stackunder[5pt]{\includegraphics[width=0.15\linewidth]{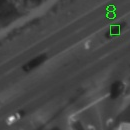}}{frame 6}
\stackunder[5pt]{\includegraphics[width=0.15\linewidth]{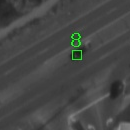}}{frame 10}
\stackunder[5pt]{\includegraphics[width=0.15\linewidth]{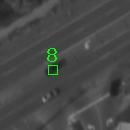}}{frame 17}
\stackunder[5pt]{\includegraphics[width=0.15\linewidth]{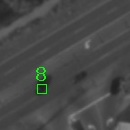}}{frame 22}
\stackunder[5pt]{\includegraphics[width=0.15\linewidth]{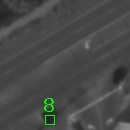}}{frame 24}
\end{center}
\caption{Snapshots of results from our two methods. First row shows the result using our DBT only. Second row shows the result by coupling DBT + LCT. Tracking results of other targets are not shown for clarity.}
\label{fig:lct}
\end{figure}

Fig. \ref{fig:lct} shows the qualitative comparison between our methods with DBT only and with DBT + LCT. The target slows down then stops to make a left turn. Motion detection fails from frame 11 to frame 21. Without LCT, the track breaks into two tracks as shown in the first row of Fig. \ref{fig:lct}, the yellow arrow indicates the missing detection at frame 17. On the contrary, DBT + LCT successfully recovers missing detections and continue the track.

Table \ref{tab:one} shows the quantitative results. By inserting additional hypotheses from potential merged blobs in the optimization, our DBT reduces the number of \textit{S/T} significantly. Combining LCT with DBT, we recover many missing detections between DBT. This increases \textit{Recall} and reduces \textit{B/T}. Note that false alarms and id-switches further decrease because LCT reduces the gap between DBT tracks and partially avoid errors from linear interpolation. Our method outperforms state-of-the-art methods in all indexes.
\begin{table*}[t]
\begin{center}
\begin{tabular}{|c|c|c|c|c|c|c|c|c|}
\hline
Method & Recall & Precision & FP/F & FP/GT & MODA & S/T & B/T & MOTA\\
\hline
Reilly \etal~\cite{Reilly:2010}
& 0.562 & 0.841 & 2.344 & 0.106 & 0.455 & 0.545 & 1.591 & 0.444\\ 
Prokaj \etal~\cite{ProkajCVPRW}
& 0.519 & 0.953 & 0.563 & 0.025 & 0.493 & 0.205 & 1 & 0.489\\
Prokaj \etal~\cite{ProkajCVPR14}
& 0.569 & 0.905 & 1.323 & 0.06 & 0.509 & 0.773 & 1.455 & 0.493\\ 
Rezatofighi \etal~\cite{jpdaR}
& 0.554 & 0.842 & 2.292 & 0.104 & 0.45 & 0.227 & 1.205 & 0.445\\
Chen \etal~\cite{ChenAVSS15}
& 0.5 & 0.973 & 0.302 & 0.014 & 0.486 & 0.023 & 0.636 & 0.486\\
\hline
Ours (DBT only) 
& 0.497 & {\bf 0.999 } & {\bf 0.01} & {\bf 0.001} & 0.497 & {\bf 0 }& 0.568 & 0.497\\
Ours (DBT + LCT)
& {\bf 0.761} & {\bf 0.999} & {\bf 0.01} & {\bf 0.001} & {\bf 0.761} & {\bf 0} & {\bf 0.159} & {\bf 0.761}\\
\hline
\end{tabular}
\end{center}
\caption{Comparison of tracking results of the Rochester sequence. Best results in each indexes are shown in bold text.}
\label{tab:two}
\end{table*}
\subsection{Results on Rochester Sequence}
We select a sequence with 650 pixels $\times$ 650 pixels (250 m $\times$ 250 m) from Rochester dataset, which is captured from the city of Rochester, NY, USA. This imagery is recorded at 2 Hz and the ground sampling distance is around 0.38 meters per pixel. We manually label ground truth for each target from the frame it starts to move to the frame right before it leaves the scene. The sequence contains 96 frames and 44 tracks. Since it is an urban-view dataset, many vehicles stop at intersections for a long period. 

The quantitive results are shown in Table \ref{tab:two}. We do not produce any swaps in this sequence in both of our methods. Using LCT further improves more than $19\%$ in \textit{recall} from move-then-stop situations. Methods that rely on only motion detections \cite{Reilly:2010,ProkajCVPRW,jpdaR,ChenAVSS15} fail in these cases. 

This sequence has higher ground sampling distance and lower contrast than the WPFAB 2009 sequence. These factors make the appearance of a target less discriminating from the background. Therefore, although \cite{ProkajCVPR14} performs the second best among all methods in \textit{recall}, it introduces many false alarms because the regression based tracker may drift to visually similar background patches. On the contrary, LCT recovers missing detections by exploring stronger evidence based on context information. We maintain high precision compared with other methods. Again, DBT + LCT is clearly the leader in all indexes.
\subsection{Computation Time}
Our approach is implemented using C++ on a desktop with 3.6GHz CPU, 16GB memory and a NVIDIA GeForce GTX 580 GPU. We use GPU only for dense flow computation using FlowLib \cite{Zach2007}. Table \ref{tab:three} shows the computational cost of methods with C++ implementation. Our DBT achieves similar computation time compared with other detection association methods \cite{Reilly:2010,ProkajCVPRW,ChenAVSS15}. 

The major limitation of our DBT + LCT is the higher computational cost compared with other DBTs \cite{Reilly:2010,ProkajCVPRW,ChenAVSS15}. In the WPFAB 2009 sequence, dense flow calculation takes 0.482 seconds which accounts for nearly half of computation time in our DBT + LCT. However, we are still more efficient than the state-of-the-art hybrid approach \cite{ProkajCVPR14}. In the Rochester sequence, since the image size is smaller, the flow computation time reduced to 0.013 sec/frame, and our DBT + LCT takes 0.361 sec/frame. Notice that the computation time of \cite{ProkajCVPR14} increases because appearance models have to update whenever regression trackers lose track. Failure of the regression tracker happens more frequently in the Rochester sequence where target appearance is less discriminating.
\begin{table*}[t]
\begin{center}
\begin{tabular}{|c|c|c|c|c|c|c|}
\hline
&Reilly &Prokaj &Prokaj &Chen&Ours&Ours\\
&\etal~\cite{Reilly:2010}&\etal~\cite{ProkajCVPRW}&\etal~\cite{ProkajCVPR14}&\etal~\cite{ChenAVSS15}&(DBT)&(DBT+LCT)\\
\hline
WPFAB & 0.121 & 0.094 & 9.413 & 0.21 & 0.124 & 1.009\\
\hline
Rochester  & 0.093 & 0.026 & 15.166 & 0.064 & 0.018 & 0.361\\
\hline
\end{tabular}
\end{center}
\caption{Comparison of computation time in seconds per frame in the WPFAB 2009 sequence and the Rochester sequence.}
\label{tab:three}
\end{table*}
\section{Conclusions}
\label{sec:conc}
Existing multi-target tracking approaches in WAMI have limitations in recovering long-term missing detections from slow or stopped targets. We propose a unified approach which couples LCT with DBT. Instead of merely relying on an appearance model, LCT explores context information between a target and its motion neighbors so that it is robust against neighboring distracters and background clutter. Hypotheses are generated by dense and sparse flow efficiently. We reduce id-switches by handling merged detections in DBT with additional hypotheses. Our DBT+LCT significantly improves tracking results compared with state-of-the-art methods on two WAMI sequences. 

Our future work is to reduce the computation time by using parallel programming. Note that both DBT and LCT can be processed in parallel for each target; therefore, we expect obvious reduction in computation time. Furthermore, we want to infer the association between tracks under long-term occlusion, which often happens in urban view scenarios.
\clearpage
\bibliographystyle{splncs}
\bibliography{egbib}
\end{document}